# Accurate Medical Named Entity Recognition Through Specialized NLP Models

Jiacheng Hu
Tulane University
New Orleans, USA

Runyuan Bao
Johns Hopkins University
Baltimore, USA

Yang Lin
University of Pennsylvania
Philadelphia, USA

Hanchao Zhang
New York University
New York, USA

Yanlin Xiang*
University of Houston
Houston, USA

*Abstract*—This study evaluated the effect of BioBERT in medical text processing for the task of medical named entity recognition. Through comparative experiments with models such as BERT, ClinicalBERT, SciBERT, and BlueBERT, the results showed that BioBERT achieved the best performance in both precision and F1 score, verifying its applicability and superiority in the medical field. BioBERT enhances its ability to understand professional terms and complex medical texts through pre-training on biomedical data, providing a powerful tool for medical information extraction and clinical decision support. The study also explored the privacy and compliance challenges of BioBERT when processing medical data, and proposed future research directions for combining other medical-specific models to improve generalization and robustness. With the development of deep learning technology, the potential of BioBERT in application fields such as intelligent medicine, personalized treatment, and disease prediction will be further expanded. Future research can focus on the real-time and interpretability of the model to promote its widespread application in the medical field.

## I. INTRODUCTION

In the modern medical field, with the popularization of data digitization and electronic medical record systems, massive amounts of unstructured text data have become an important resource for medical research and clinical applications [1]. Electronic medical records, research papers, drug instructions, and diagnostic records provide comprehensive information about patients, disease records, and treatment plans, serving as valuable references for medical institutions and researchers [2]. However, faced with such a large amount of data and complex medical terminology, traditional data analysis methods face huge challenges in processing these unstructured data [3]. Therefore, natural language processing (NLP) technology has gradually become an indispensable tool in medical data processing, and named entity recognition (NER), as an important branch of NLP, is particularly critical in applications such as medical information extraction and disease prediction [4]. How to accurately extract key information such as diseases, drugs, and symptoms from complex medical texts is crucial to improving the efficiency of medical services and the intelligence level of medical information systems.

The BioBERT (Bidirectional Encoder Representations from Transformers for Biomedical Text Mining) model is a pre-trained model specifically for the biomedical field based on the classic BERT (bidirectional encoder representation) [5]. By pre-training on a large-scale biomedical corpus, BioBERT can more accurately understand the medical field's proper nouns, professional terms, and complex contextual relationships. Compared with traditional general NLP models, BioBERT performs better in medical named entity recognition (such as disease, drug, symptom recognition) and can process text details in various medical scenarios. With its deep semantic understanding ability in the biomedical field, BioBERT can not only effectively extract key information from medical texts, but also provide important support for subsequent clinical decision-making, disease monitoring, info recommendations [6], etc., greatly promoting the intelligent development of medical information processing.

At present, in the application of medical named entity recognition, accuracy, precision, and efficiency are always the main concerns [7]. The addition of BioBERT not only significantly improves the accuracy of the recognition results, but also effectively reduces the misrecognition rate by eliminating ambiguous information and reducing erroneous labels [8]. Its advantage lies in the deep learning and precise understanding of the medical field's proprietary vocabulary and terminology, which makes BioBERT have important application potential in actual medical scenarios such as disease prediction, personalized treatment plan recommendation, and patient management. At the same time, with the rapid development of deep learning technology, BioBERT has also performed well in the processing ability of long texts and the accurate grasp of contextual relationships [9]. Specifically, BioBERT is crucial in the Human-Computer Interaction (HCI) field [10], as it enhances intelligent health systems, allowing for natural, context-aware interaction between users and healthcare technologies. In the Emotion-Aware Interaction field [11], BioBERT improves personalized interfaces by understanding user inputs related to medical queries, enabling more effective health monitoring and educational tools, providing solid technical support for intelligent data processing in the medical field, and laying the foundation for the further development of medical information systems in the future.

However, the privacy protection of medical data also comes with it. Medical information is highly sensitive, and data leakage may have serious consequences. In this context, BioBERT can achieve information extraction tasks while effectively protecting patient privacy through specific data processing methods [12]. It uses data de-identification and secure computing technology to convert raw data into text vectors that cannot be directly identified, reducing the risk of patient privacy leakage. This privacy protection feature enables BioBERT to comply with strict medical data security specifications and privacy protection requirements during research and application. With the continuous improvement of global awareness of data security and privacy protection, the application of BioBERT in the field of medical data processing meets social and legal needs and is expected to become a basic tool for widespread use in medical data processing in the future.

Medical named entity recognition using BioBERT not only enhances medical data analysis but also provides essential technical support for the development of intelligent medicine [13]. As an optimized deep learning model, BioBERT significantly improves the efficiency and accuracy of medical text processing. With the continuous advancement of natural language processing (NLP) technology and deep learning models, BioBERT holds great promise for future research and applications, offering strong support for intelligent and precision medicine [14]. Its application will have a profound impact on areas such as disease management, health monitoring, and personalized treatment. By further optimizing BioBERT's use in medical named entity recognition, we move towards a new era of data-driven, intelligent healthcare.

## II. METHOD

In implementing medical named entity recognition based on BioBERT, the biomedical text data is first preprocessed and converted into a format acceptable to the BioBERT model. The BioBERT model encodes text features through a bidirectional Transformer structure. The input is a sequence of medical text segmentations. After being processed by multiple layers of Transformers, the feature vector representation of the text is output. Its network structure is shown in Figure 1.

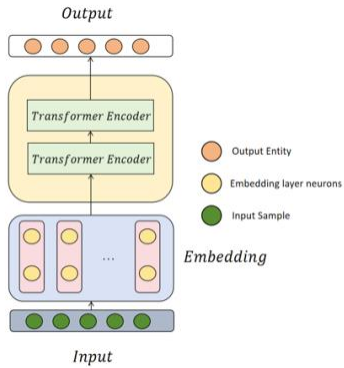

Figure 1 Article network architecture diagram

The feature vector of each input word is used to predict whether it belongs to a named entity [15]. The loss function of the model uses cross-entropy loss, and its formula is:

$$L = -\frac{1}{N}\sum_{i=1}^{N}[y_i \log(y_i') + (1 - y_i)\log(1 - y_i')] \quad (1)$$

Among them, A represents the true label, B represents the predicted probability, and C represents the number of samples. Through this loss function, the model can better fit the medical named entity recognition task.

Secondly, in the feature representation stage, BioBERT effectively captures the contextual dependencies of the text through the self-attention mechanism [16] to provide a more accurate semantic understanding of medical terms. The calculation of its self-attention mechanism is as follows:

$$\text{Attention}(Q, K, V) = \text{softmax}\left(\frac{QK^\top}{\sqrt{d_k}}\right)V \quad (2)$$

Among them, Q, K, and V represent query, key, and value vector matrices respectively, and $\sqrt{d_k}$ is the dimension of the vector. Through self-attention, the model can focus on important words and contextual relationships in the text, thereby improving the semantic representation ability of medical text.

Finally, in order to realize the recognition of named entities, the feature vector of each word is input into the fully connected layer, and then the probability of each category is normalized through the Softmax activation function, and the prediction result is the entity type to which the word belongs. The Softmax function is defined as follows:

$$\text{Softmax}(z_i) = \frac{e^{z_i}}{\sum_j e^{z_j}} \quad (3)$$

Among them, $z_i$ represents the score of the i-th category, and the probability distribution of each category is obtained after Softmax normalization. During the training process, the model continuously adjusts parameters to achieve the best performance on the validation set to ensure accuracy in the named entity recognition task.

## III. EXPERIMENT

### A. Datasets

In order to conduct the medical named entity recognition experiment based on BioBERT, this study used the MIMIC-III dataset. The MIMIC-III (Medical Information Mart for Intensive Care III) dataset was collected by the Beth Israel Deaconess Medical Center of Harvard University and contains detailed medical records of more than 40,000 patients in the intensive care unit (ICU). This dataset is widely used in medical text mining and medical named entity recognition tasks, with rich text content and diverse patient information, including diagnosis records, drug use, laboratory test results, etc.

The text records in the MIMIC-III dataset contain a large number of medical terms and disease descriptions, making it an ideal data source for medical named entity recognition tasks. In the data preprocessing stage, the text content is first de-identified to protect patient privacy; then, the text is segmented and annotated to adapt to the input format of the BioBERT model. In addition, each record in the dataset is attached with corresponding labels, including disease names, drugs, treatment plans, etc., to facilitate supervised learning of the model.

In model training, the MIMIC-III dataset can be used to effectively evaluate the performance of BioBERT in identifying medical-named entities. By testing on real medical record data, the model can learn specific entity patterns in medical texts and improve the accuracy of named entity recognition. The diversity and authenticity of the dataset ensure the wide applicability of the experimental results and provide a solid data foundation for subsequent medical information processing and decision support.

*B. Experimental setup*

In the experimental setting, first, we divided the MIMIC-III dataset into training set, validation set and test set, accounting for 70%, 15% and 15% respectively, to ensure that the model has good generalization ability on unseen data. During the division process, the integrity of each record is retained to maintain the semantic and contextual information of the medical text. At the same time, each text record is segmented and annotated to adapt to the input format of the BioBERT model and ensure the accurate annotation of medical named entities.

Secondly, the BioBERT model in the experiment loads the parameters pre-trained by biomedicine in the initial stage to converge faster. During the training process, the Adam optimizer is used, the initial learning rate is set to 2e-5, and a dynamic learning rate decay strategy is adopted to automatically adjust the learning rate according to the performance of the validation set to avoid overfitting problems. The batch size is set to 16 to balance computational efficiency and model effect. In addition, during the training process, the cross-entropy loss function is used to measure the performance of the model on the named entity recognition task to ensure that the model can accurately identify the entity types in the medical text.

Finally, in order to evaluate the named entity recognition effect of the model on the test set, precision, recall and F1 score were used as the main evaluation indicators. These indicators can fully reflect the recognition ability of the model and its performance in real medical texts. By analyzing the experimental results, we can gain an in-depth understanding of the advantages and disadvantages of BioBERT in the medical named entity recognition task, thus providing a reference for future improvements and optimizations.

*C. Experiments Comparation*

In the experiment of medical named entity recognition, in order to better evaluate the effect of BioBERT, we introduced multiple comparison models, including BERT, ClinicalBERT, SciBERT, BlueBERT and RoBERTa. These models are pre-trained on different data and domains and have their own characteristics [17-19]. As a general language model, BERT is widely used in various NLP tasks, and its bidirectional Transformer structure can capture the bidirectional contextual relationship of text. Although BERT performs well in medical tasks, it lacks medical-specific pre-training corpus and may not be as good as BioBERT in recognizing professional medical terms [20].

ClinicalBERT is a version of BERT fine-tuned specifically on clinical texts, suitable for clinical scenarios such as electronic health records (EHR). It has a better understanding of clinical-specific terms and helps improve the named entity recognition effect of medical texts [21]. SciBERT focuses on scientific literature, especially literature in the fields of biomedicine and computer science, which gives it an advantage in academic papers and research data analysis [22]. Although SciBERT's pre-training corpus involves the biomedical field, there is still a gap with real clinical data, so its performance in real medical texts may not be as good as ClinicalBERT and BioBERT.

BlueBERT is a model that combines PubMed and MIMIC-III datasets. By pre-training on scientific literature and medical records, BlueBERT can better adapt to texts containing professional terms and patient data. Therefore, it has a strong performance in named entity recognition tasks for medical and clinical texts. RoBERTa is an optimized version of BERT. It performs well on general NLP tasks through pre-training with larger data volumes and longer training time. Although RoBERTa is not specialized in the medical field, it is still competitive in medical named entity recognition tasks with sufficient data and further fine-tuning. The experimental results are shown in Table 1.

Table 1 Experiment result

| Model | Precision% | F1-score |
|---|---|---|
| Bert [23] | 82.5 | 81.0 |
| ClinicalBERT [24] | 85.2 | 83.5 |
| SciBert [25] | 84.1 | 82.8 |
| BlueBert [26] | 87.3 | 85.0 |
| BioBert [27] | 89.8 | 87.6 |

Through the experimental results, we can see that BioBERT performs best in the task of medical named entity recognition, surpassing other models in precision and F1 score, reaching 89.8% and 87.6% respectively. This result shows that BioBERT can more effectively identify named entities in medical texts, such as diseases, drugs, and symptoms, and its performance improvement is mainly due to its pre-training in the biomedical field. Since BioBERT has been fine-tuned on a large amount of medical corpus, the model has stronger domain adaptability and can accurately understand complex terms and professional expressions in medical texts.

In contrast, BlueBERT's precision and F1 score are second only to BioBERT, at 87.3% and 85.0% respectively, indicating that it has a good performance in the task of medical named entity recognition. BlueBERT is a model pre-trained based on PubMed and MIMIC-III datasets. By combining biomedical literature and clinical record data, BlueBERT can effectively capture specific contextual information in the medical field. In the experiment, BlueBERT outperformed General BERT and SciBERT, but still slightly behind BioBERT. This may be because BlueBERT's pre-training data, while covering a wide range, is not as refined as BioBERT in the biomedical field, especially in the fine-grained recognition of named entities.

ClinicalBERT's precision and F1 score are 85.2% and 83.5%, respectively, which are in the upper-middle level for this task. Although ClinicalBERT has advantages in handling specialized clinical terms, it lacks training data for a wide range of medical literature, and thus is slightly inferior to BioBERT and BlueBERT in generalization ability. This makes it difficult for ClinicalBERT to match BioBERT's high precision and F1 score when faced with complex medical expressions and more detailed named entities. However, for tasks focused on electronic health records, ClinicalBERT's performance is still competitive.

SciBERT has a precision and F1 score of 84.1% and 82.8%, which is slightly better than the general BERT, but lower than the domain-specific BioBERT, BlueBERT, and ClinicalBERT. SciBERT is mainly pre-trained based on literature in the scientific field and performs relatively well in biomedical literature. However, its training data is not limited to the medical field but also involves other disciplines, so SciBERT's performance in medical named entity recognition tasks is slightly limited. This results in SciBERT not being able to deeply understand the complex contextual relationships in medical texts like BioBERT and BlueBERT. Therefore, although SciBERT performs well in academic research, it does not perform as well as other medical-specific models in specific medical tasks.

The precision and F1 scores of general BERT are 82.5% and 81.0%, respectively, which is the weakest among all models. As a general NLP model, although BERT has a powerful bidirectional Transformer structure that can capture contextual relationships in general texts, it lacks specialized pre-training in the medical field and cannot accurately understand the professional terms and complex contexts in the medical field. Therefore, in the task of medical named entity recognition, BERT's performance lags significantly behind other models fine-tuned on biomedical or clinical texts. This shows that for medical named entity recognition, the pre-training advantages of domain-specific models such as BioBERT and BlueBERT are very obvious, effectively improving the adaptability and accuracy of the model.

In summary, the experimental results show that BioBERT outperforms other models in terms of precision and F1 score, verifying its applicability and superior performance in the task of medical named entity recognition.

Finally, we also give the loss function decline graph during training, as shown in Figure 2.

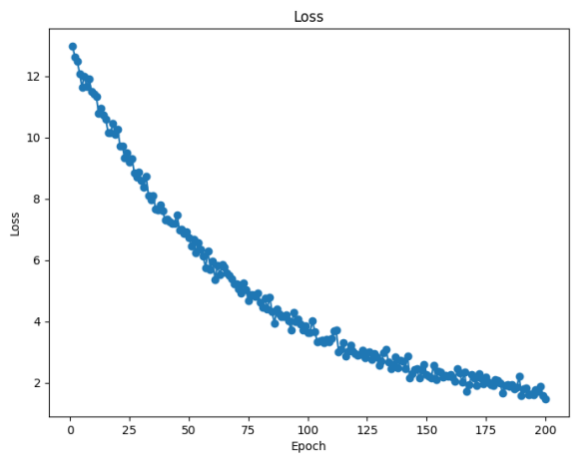

Figure 2 The loss function decreases with epoch

This figure shows the change of the loss of the model during the training process. The horizontal axis represents the number of training epochs, and the vertical axis represents the loss value. It can be seen that with the increase of training epochs, the loss value shows a clear downward trend. This trend shows that the model is constantly learning and gradually optimizing parameters so that it fits the training data better. The loss value is high in the initial stage, and the model error is large, but as the training progresses, the loss decreases rapidly.

After about 50 epochs, the rate of decrease of the loss value begins to slow down, but it is still steadily decreasing overall, indicating that the model is gradually converging. Although the loss value fluctuates slightly in individual epochs, the overall trend is still downward, indicating that the training effect of the model continues to improve. These fluctuations may be caused by the randomness of batch data and do not affect the overall convergence of the model.

Finally, when it is close to 200 epochs, the loss value tends to stabilize, indicating that the model has basically reached the optimal state and no longer shows a significant error decrease. This stable state indicates that the model training process has been completed or is nearly completed. The convergence of the graph reflects that the training quality of the model is good. The loss gradually decreases and finally stabilizes, which means that the learning effect of the model is relatively ideal.

## IV. Conclusion

Through this study, we verified the superiority of BioBERT in the medical named entity recognition task. Experimental results show that BioBERT performs best in accuracy and F1 score, significantly surpassing other models. This result demonstrates the adaptability and accuracy of BioBERT in the medical field. Especially when dealing with complex medical terminology and professional texts, BioBERT can better extract key information, providing powerful tool support for clinical data analysis and medical decision support.

However, this study also has some limitations. Although BioBERT performs superiorly in specific medical named entity recognition tasks, its performance in other tasks or specific medical subfields still requires further verification. In addition, the privacy and compliance requirements of medical data also pose challenges to the practical application of BioBERT. In future research, the integration of BioBERT with other medical-specific models can be further explored to improve the model's generalization ability and robustness to uncertain data to better cope with diverse clinical needs.

Looking to the future, with the continuous development of deep learning and natural language processing technology, BioBERT is expected to be applied in more medical scenarios. Combining the latest pre-training technology with larger biomedical data sets, BioBERT and its improved versions may achieve better results in medical text analysis, disease prediction, personalized medicine, and other fields. Future research can also focus on the real-time and interpretability of models to promote the popularization and improvement of intelligent medical systems and help achieve the long-term goals of precision medicine and smart medicine.